\documentclass{article}
\usepackage{arxiv}
\usepackage[T1]{fontenc}
\usepackage[utf8]{inputenc}
\usepackage{microtype}
\usepackage{amsmath,amssymb,amsfonts}
\usepackage{booktabs}
\usepackage{graphicx}
\usepackage{array}
\usepackage{multirow}
\usepackage{natbib}
\usepackage{hyperref}
\usepackage{url}
\usepackage{cleveref}
\usepackage{doi}

\title{Cyber-Physical RSI for AI-RAN: A Case Study}
\author{}
\date{}

\renewcommand{\shorttitle}{Cyber-Physical RSI for AI-RAN}
\renewcommand{\headeright}{Preprint}
\renewcommand{\undertitle}{Preprint}
\hypersetup{
  colorlinks=true, linkcolor=black, citecolor=black, urlcolor=blue,
  pdftitle={Cyber-Physical RSI for AI-RAN: A Case Study},
  pdfsubject={Cyber-physical recursive self-improvement for AI-RAN},
  pdfkeywords={AI-RAN, recursive self-improvement, cyber-physical systems, handover, physical feedback}
}
\graphicspath{{figures/}}

\author{{Binghan Wu} \\
	AsiaInfo Technologies\\
	Beijing, China\\
	\texttt{wubh3@asiainfo.com} \\
	\And
	{Kai Zeng} \\
	AsiaInfo Technologies\\
	Beijing, China\\
	\texttt{zengkai@asiainfo.com}\\
  \And
	{Zhiqi Li} \\
	AsiaInfo Technologies\\
	Beijing, China\\
	\texttt{lizq18@asiainfo.com}\\
  \And
	{Zhigang Wang} \\
	AsiaInfo Technologies\\
	Beijing, China\\
	\texttt{wangzg6@asiainfo.com}\\
	\And
	{Yang Liu} \\
	Hong Kong Polytechnic University\\
	Hong Kong, China\\
	\texttt{yang-veronica.liu@polyu.edu.hk}\\
	\And
	{Ya-Qin Zhang} \\
	Institute for AI Industry Research (AIR)\\
	Tsinghua University\\
	Beijing, China\\
	\texttt{zhangyaqin@air.tsinghua.edu.cn}\\
	\And
	{Joseph Sifakis} \\
	Verimag\\
	Université Grenoble Alpes\\
	Grenoble, France\\
	\texttt{joseph.sifakis@univ-grenoble-alpes.fr}\\
	\And
	{Ye Ouyang} \\
	Institute for AI Industry Research (AIR)\\ Tsinghua University\\
	Beijing, China\\
	\texttt{ouyangye@air.tsinghua.edu.cn}\\
}

\begin{document}
\maketitle

\begin{abstract}
Learned models for radio access networks (RANs) are often developed using simulation, yet their predictions can shift when exposed to physical propagation conditions, protocol behavior, and equipment implementations. This paper presents a cyber-physical recursive self-improvement (RSI) framework for AI-RAN that connects a task-oriented student model, an experiment-planning coach agent, a calibratable virtual environment, and a physical host agent. Physical measurements are organized as evidence for both model adaptation and subsequent experiment planning. We examine an implementation of the physical-feedback pathway using ping-pong handover as a case study. The system completes 12 physical executions organized into six baseline--candidate pairs, updates the student from measured outcomes, and evaluates the revised model on reserved physical configurations. In the observed short-window tests, increasing the time-to-trigger from 320 to 1280\,ms changes the ping-pong count from three to zero. The updated student achieves a grouped-validation mean absolute error (MAE) of 2.673 round trips/min, while a common three-configuration physical evaluation yields MAEs of 1.240 and 1.601 round trips/min for the pre-feedback and post-feedback models, respectively. These results characterize both the operational feedback loop and the parameter-dependent generalization behavior revealed by new physical experiments.
\end{abstract}

\keywords{AI-RAN \and recursive self-improvement \and cyber-physical systems \and autonomous experimentation \and mobility management}

\section{Introduction}
\label{sec:intro}

The integration of machine learning into radio access networks (RANs) creates opportunities to move from offline network analysis toward continuously updated prediction and control. AI-RAN explores how wireless communication and AI computing can coexist on a common infrastructure~\citep{kundu2026airan}, while O-RAN provides open interfaces and architectural support for intelligent network operations~\citep{polese2023understanding}. On the development side, programmable testbeds and RAN digital twins can reproduce radio conditions, generate training data, and compare configurations before physical deployment~\citep{polese2024colosseum,huang2026digitaltwin}. These developments raise a practical question: how can models learned in virtual environments be adapted using evidence from physical wireless systems?

A central obstacle is the discrepancy between simulated and observed network behavior. Radio propagation and mobility are approximated in virtual environments, whereas physical outcomes additionally depend on device implementations, protocol procedures, and measurement conditions. The resulting gap affects not only the predictions of a learned task model but also the value of experiments selected using that model. Physical measurements help resolve these discrepancies, although configuring and resetting wireless equipment makes each experiment more costly than a virtual run. Model adaptation and experiment selection therefore become coupled problems: the system must decide both \emph{what to learn from new physical observations} and \emph{which physical observations are worth acquiring next}.

Mobility management offers a concrete setting for studying this coupling. A ping-pong handover occurs when a user equipment (UE) switches back and forth between neighboring cells within a short period. Event-A3-related offsets, hysteresis, and time-to-trigger (TTT) affect the timing and conditions of handover. A configuration that suppresses a short-lived trigger may reduce unnecessary handovers, whereas the same choice can change responsiveness in other mobility conditions. Learning configuration effects consequently requires evidence about actual network responses, not only a simulated ranking of parameters.

Active learning addresses how to select informative observations~\citep{settles2011theories}, and automated physical experimentation connects experiment selection to instrumentation and measurement~\citep{shields2023active}. Recursive self-improvement (RSI) adds another dimension: experience can be used to revise the process that generates subsequent improvements. Work on self-improving software agents illustrates feedback-guided changes to agent implementations~\citep{zhang2026darwin}; recent work on physical RSI examines how this principle interacts with embodied observations, experiment costs, and the persistence of learned improvements~\citep{luo2026physicalrsi}. For AI-RAN, a useful realization is to separate the model performing a network task from the agent organizing its improvement.

We develop a \emph{Cyber-Physical RSI} framework around this separation. A \emph{student} learns network-task predictions, while a \emph{coach} selects experiments, coordinates training and assessment, and maintains experimental memory. A physical host agent translates structured experiment requests into bounded device operations, and a virtual environment provides a complementary medium for scenario exploration and alignment with physical evidence. Rather than placing a large language model in the real-time handover control path, the coach operates at the experimental time scale.

This paper makes three contributions. First, it specifies a cyber-physical architecture that distinguishes student adaptation from feedback-driven experiment planning. Second, it implements the physical-experiment pathway, including execution, measurement provenance, dataset updates, and model assessment. Third, it reports paired handover interventions and a common physical evaluation of student versions before and after feedback, exposing configuration-specific differences that can guide subsequent experiments.

\section{Cyber-Physical RSI Framework}
\label{sec:framework}

\subsection{Student and Coach: Two Coupled Improvement Processes}

Cyber-Physical RSI links two time scales. At the task time scale, the student $M_k$ maps wireless observations and configuration conditions to an event judgment or configuration-effect prediction. At the experimental time scale, the coach analyzes model performance, selects new conditions to measure, and coordinates dataset updates, training, and assessment. A physical experiment consequently serves two purposes: it supplies training observations to the student and produces evidence that can inform the coach's later choices.

Let $S_k$ denote the virtual environment and $H_k$ the coach's versioned experimental memory at improvement round $k$. For an experiment $e_k$ producing verified physical evidence $D_k^{\mathrm P}$, the joint evolution of the framework is represented as
\begin{equation}
\label{eq:update}
(S_k,M_k,H_k) \xrightarrow{e_k,\,D_k^{\mathrm P}}
(S_{k+1},M_{k+1},H_{k+1}).
\end{equation}
Here, updating $M_k$ corresponds to task-model adaptation; updating $S_k$ aligns the virtual environment with observed physical behavior; and updating $H_k$ retains the experiment's configuration, outcome, and model-assessment evidence. The coach's underlying language model can remain fixed while its decisions change through updated memory, tools, and environment models.

Because physical trials are costly, experiment selection can be expressed as a budget-constrained problem:
\begin{equation}
\label{eq:budget}
\min_{\Pi}\;\mathbb{E}\bigl[J_{\mathrm P}(M_K)\bigr]
\quad\text{subject to}\quad
\sum_{k=0}^{K-1}c_{\mathrm P}(e_k)\leq B,
\end{equation}
where $\Pi$ is the experiment-selection and model-update procedure, $c_{\mathrm P}$ is physical experiment cost, and $J_{\mathrm P}$ measures performance on physical network tasks. An operational handover objective can combine ping-pong events, handover reliability, radio-link failures, and service quality. The student tasks evaluated in this case study are observed-event classification and prediction of ping-pong frequency under a candidate configuration.

\subsection{Architecture and Interfaces}
\label{subsec:architecture}

\begin{figure}[t]
  \centering
  \includegraphics[width=\linewidth]{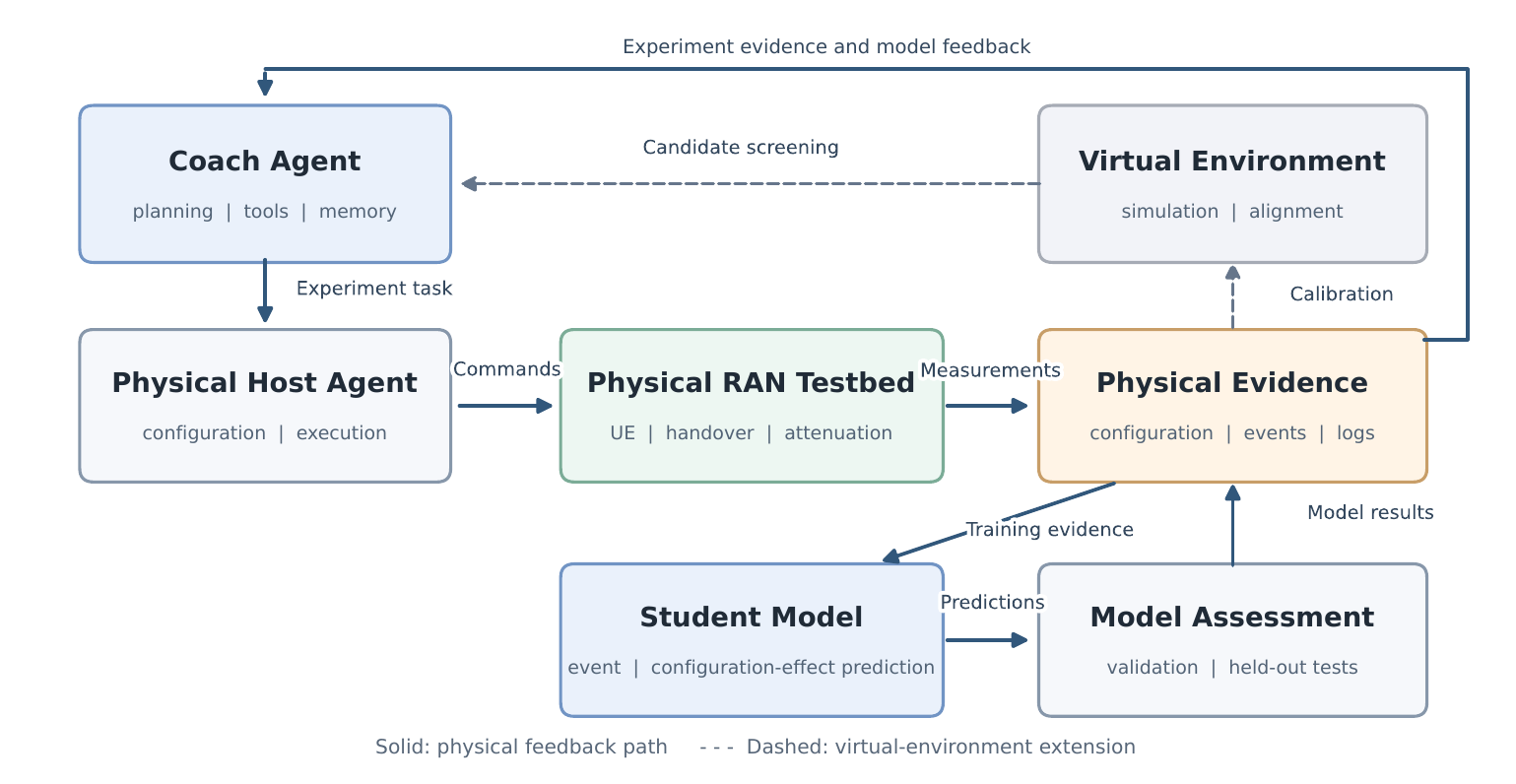}
  \caption{Cyber-Physical RSI architecture. Solid arrows indicate the implemented physical-experiment and student-feedback pathway; dashed arrows show the virtual-environment alignment and candidate-screening extension. Experimental evidence also enters the coach's versioned memory for subsequent planning.}
  \label{fig:architecture}
\end{figure}

\Cref{fig:architecture} presents the main components and their information flow.

\paragraph{Physical environment and host agent.}
The physical side consists of radio equipment, a UE, programmable attenuation, and measurement facilities. The host agent receives a structured request specifying experimental conditions and measurement requirements. It performs configuration, execution, logging, and state restoration through supported device operations. The resulting evidence record links the realized configuration to the observation window and measured events. This interface separates experiment planning from low-level equipment control.

\paragraph{Virtual environment.}
The virtual environment provides repeatable, lower-cost representations of radio conditions for model development and candidate screening. Physical evidence can be used to refine parameters related to propagation, attenuation, and handover behavior. The relevant alignment targets are decision-related responses, including whether an event occurs and how candidate configurations compare, rather than signal traces alone.

\paragraph{Student.}
The student learns relationships between radio observations, configuration parameters, and measured outcomes. In the present handover case, its two tasks are to estimate the presence of an observed ping-pong event and to predict the frequency of ping-pong round trips for candidate settings. Configuration-effect predictions can subsequently be used by a separate decision policy to compare possible settings.

\paragraph{Coach.}
The coach combines language-model reasoning with experiment, training, and evaluation tools. It constructs experiment requests, invokes the host agent, organizes the resulting evidence, and maintains model and dataset versions. Its memory $H_k$ associates attempted conditions, physical results, prediction errors, and execution outcomes with the decisions that produced them, providing a basis for subsequent experiment selection.

\subsection{Feedback Protocol}
\label{subsec:protocol}

An experimental round proceeds from planning to physical execution, evidence processing, model adaptation, and assessment. The coach identifies a condition of interest and issues a bounded experiment request. After the host agent executes the requested configuration, it returns the realized settings, measurements, and restoration record. The observations are organized into event labels and configuration-effect samples; model-update procedures then produce candidate student versions. A reserved evaluation protocol measures their predictions on physical conditions not used for the update.

Within the full framework, the same evidence can also inform the virtual environment. Calibrated virtual experiments can then screen candidates before allocating further physical trials. Both the model assessment and the discrepancies between predicted and observed behavior are retained in the coach's memory. This makes the experimental outcome useful beyond the current student update: it supplies a concrete basis for deciding which parameter region to explore next. The case study below implements and evaluates the physical-execution, student-adaptation, and physical-assessment pathway.

\section{Physical Experiment and Evaluation Design}
\label{sec:setup}

\subsection{Handover Conditions and Paired Interventions}

We study ping-pong behavior under controlled short-term radio attenuation. The prescribed signal-weakening interval lasts approximately 1.2\,s. Three handover parameters are varied: A3 offset, hysteresis, and TTT. The common baseline is an A3 offset of 1\,dB, hysteresis of 1\,dB, and TTT of 320\,ms.

A paired single-variable design is used. For each candidate setting, a baseline experiment is run, followed by an experiment that changes only one of the three parameters. Each pair therefore has its own measured baseline rather than relying on a baseline observation reused across all comparisons. The study comprises 12 physical executions, forming six baseline--candidate pairs. Each execution yields a realized configuration, the event count within a scoring window, and the completion and restoration status.

\subsection{Feedback and Reserved Physical Assessment}

The six pairs are divided into two blocks. The \emph{learning block} uses A3 offset $=2$\,dB, hysteresis $=2$\,dB, and TTT $=512$\,ms, each relative to its paired baseline. The measured outcomes contribute to the feedback dataset. The \emph{evaluation block} uses A3 offset $=3$\,dB, hysteresis $=3$\,dB, and TTT $=1280$\,ms, with its physical measurements reserved for assessing predictions at new parameter magnitudes.

The feedback update increases the number of training observations for event diagnosis from 52 to 64 and for configuration-effect prediction from 13 to 16. The corresponding numbers of session groups rise from 10 to 11 and from five to six. These are curated learning samples rather than independent experiment counts. Several student versions are generated and assessed using session-grouped validation. We use V2 as the pre-feedback comparator and V6 as the selected post-feedback model; both are evaluated against the same three reserved physical candidate configurations.

\subsection{Metrics}

For $N$ ping-pong round trips observed during an effective window of $T$ seconds, the normalized frequency is
\begin{equation}
\label{eq:rate}
r=\frac{60N}{T}\quad\text{(round trips/min)}.
\end{equation}
The effective scoring windows in these experiments are approximately 16.8\,s. The event-classification task uses the Brier score for probabilistic predictions, whereas configuration-effect prediction is assessed by mean absolute error (MAE). Historical validation averages the losses of held-out session groups. The reserved physical assessment computes MAE and root mean squared error (RMSE) over the three candidate configurations, with each configuration receiving equal weight.

\section{Experimental Results}
\label{sec:results}

\subsection{Physical Execution and Feedback Integration}

All 12 physical executions returned completed measurements and confirmed restoration of both the A3 configuration and the radio attenuation matrix. The host agent thereby connected a structured experiment request to an observable physical outcome with a documented end state. The learning-block results were incorporated into the revised dataset, which was subsequently used to generate and evaluate student candidates. The evaluation block provided common physical observations for comparing predictions from before and after the feedback update.

\subsection{Observed Response to Handover Parameters}

\begin{table}[t]
\centering
\caption{Physical ping-pong response in six baseline--candidate pairs. The common baseline is A3 offset $=1$\,dB, hysteresis $=1$\,dB, and TTT $=320$\,ms. Frequencies are in round trips/min.}
\label{tab:pairs}
\small
\setlength{\tabcolsep}{5pt}
\begin{tabular}{@{}llr r r r r@{}}
\toprule
Block & Varied parameter & Candidate & $N_\mathrm{base}$ & $N_\mathrm{cand}$ & $r_\mathrm{base}$ & $r_\mathrm{cand}$\\
\midrule
Learning & A3 offset & 2\,dB & 2 & 3 & 7.130 & 10.696\\
Learning & Hysteresis & 2\,dB & 3 & 3 & 10.680 & 10.696\\
Learning & TTT & 512\,ms & 3 & 1 & 10.692 & 3.563\\
Evaluation & A3 offset & 3\,dB & 2 & 3 & 7.122 & 10.685\\
Evaluation & Hysteresis & 3\,dB & 2 & 3 & 7.116 & 10.686\\
Evaluation & TTT & 1280\,ms & 3 & 0 & 10.677 & 0.000\\
\bottomrule
\end{tabular}
\end{table}

\begin{figure}[t]
  \centering
  \includegraphics[width=0.94\linewidth]{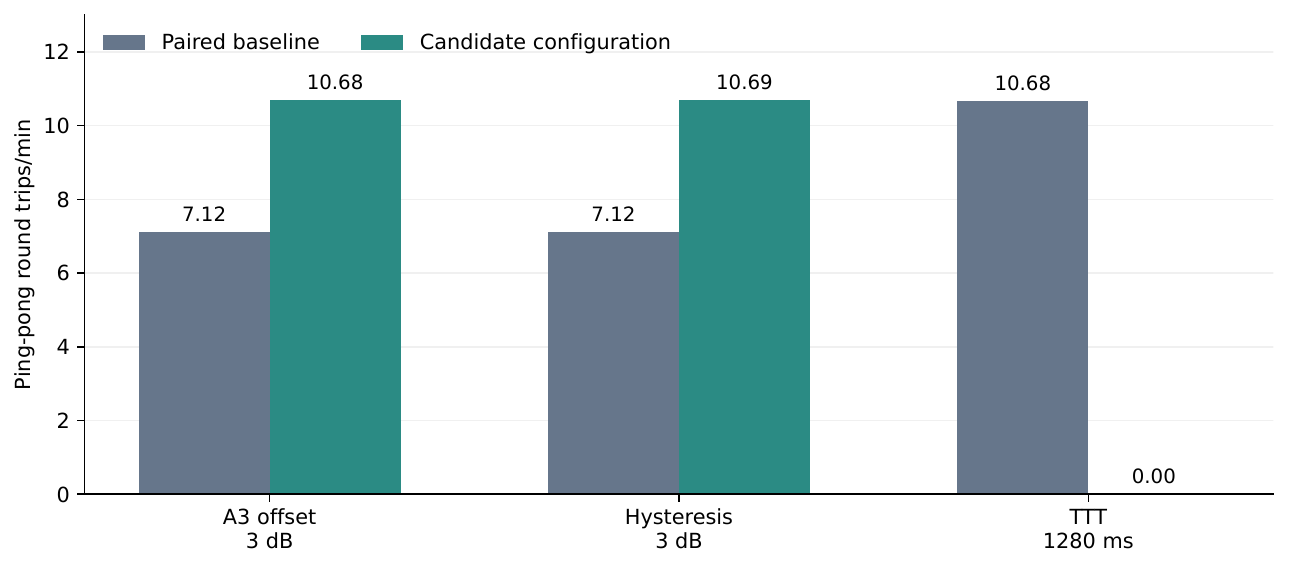}
  \caption{Observed ping-pong frequency for the three evaluation candidates and their independently executed paired baselines. Each bar represents one scoring window of approximately 16.8\,s.}
  \label{fig:physical}
\end{figure}

\Cref{tab:pairs} reports the responses from all six pairs; \cref{fig:physical} visualizes the reserved evaluation block. Increasing the A3 offset or hysteresis did not reduce the ping-pong count within the observed windows. In contrast, extending TTT from the 320\,ms baseline to 512\,ms changed the count from three to one in the learning block, and extending it to 1280\,ms changed the count from three to zero in the evaluation block.

The TTT response is consistent with the timing of a transient trigger: a condition that persists for less than the required trigger duration may expire before a handover is initiated. Under the approximately 1.2\,s attenuation interval used here, a 1280\,ms TTT provides a longer temporal filter than the transient duration. These measured outcomes distinguish the physical effects of the candidate parameters and provide informative targets for configuration-effect modeling.

\subsection{Grouped Validation After Feedback}

\begin{table}[t]
\centering
\caption{Session-grouped validation of student candidates. V2 uses the pre-feedback dataset, whereas V3--V6 use the updated dataset. The effect-prediction MAE is in round trips/min.}
\label{tab:validation}
\small
\begin{tabular}{@{}l l r r@{}}
\toprule
Model & Dataset & Event Brier $\downarrow$ & Effect MAE $\downarrow$\\
\midrule
V2 & Before feedback & 0.0000806 & 3.1942\\
V3 & After feedback & 0.0000819 & 2.8839\\
V4 & After feedback & 0.0000819 & 3.5832\\
V5 & After feedback & 0.0000819 & 3.6369\\
\textbf{V6} & \textbf{After feedback} & \textbf{0.0000819} & \textbf{2.6730}\\
\bottomrule
\end{tabular}
\end{table}

Feedback adds 12 event-diagnosis and three configuration-effect training samples. As shown in \cref{tab:validation}, V6 achieves the lowest configuration-effect MAE (2.673 round trips/min) among the four updated-data candidates. The reduction relative to V3, which uses the same dataset, is approximately 7.3\%. V4 and V5 have larger errors, reflecting differences among the candidate model updates. The session-grouped assessment is used to select V6 for evaluation on the reserved physical conditions.

\subsection{Predictions on New Physical Configurations}

\begin{table}[t]
\centering
\caption{Predicted and measured ping-pong frequencies for the common reserved physical evaluation block (round trips/min). MAE and RMSE give equal weight to each candidate.}
\label{tab:prediction}
\small
\setlength{\tabcolsep}{4pt}
\begin{tabular}{@{}l r r r r r@{}}
\toprule
Candidate & Observed & V2 pred. & V6 pred. & V2 $|e|$ & V6 $|e|$\\
\midrule
Offset $=3$\,dB & 10.6848 & 11.0651 & 11.7037 & 0.3804 & 1.0189\\
Hyst. $=3$\,dB & 10.6859 & 11.0594 & 11.6997 & 0.3735 & 1.0138\\
TTT $=1280$\,ms & 0.0000 & 2.9664 & 2.7715 & 2.9664 & 2.7715\\
\midrule
\textbf{MAE} & --- & --- & --- & \textbf{1.2401} & \textbf{1.6014}\\
\textbf{RMSE} & --- & --- & --- & \textbf{1.7401} & \textbf{1.8025}\\
\bottomrule
\end{tabular}
\end{table}

\begin{figure}[t]
  \centering
  \includegraphics[width=0.94\linewidth]{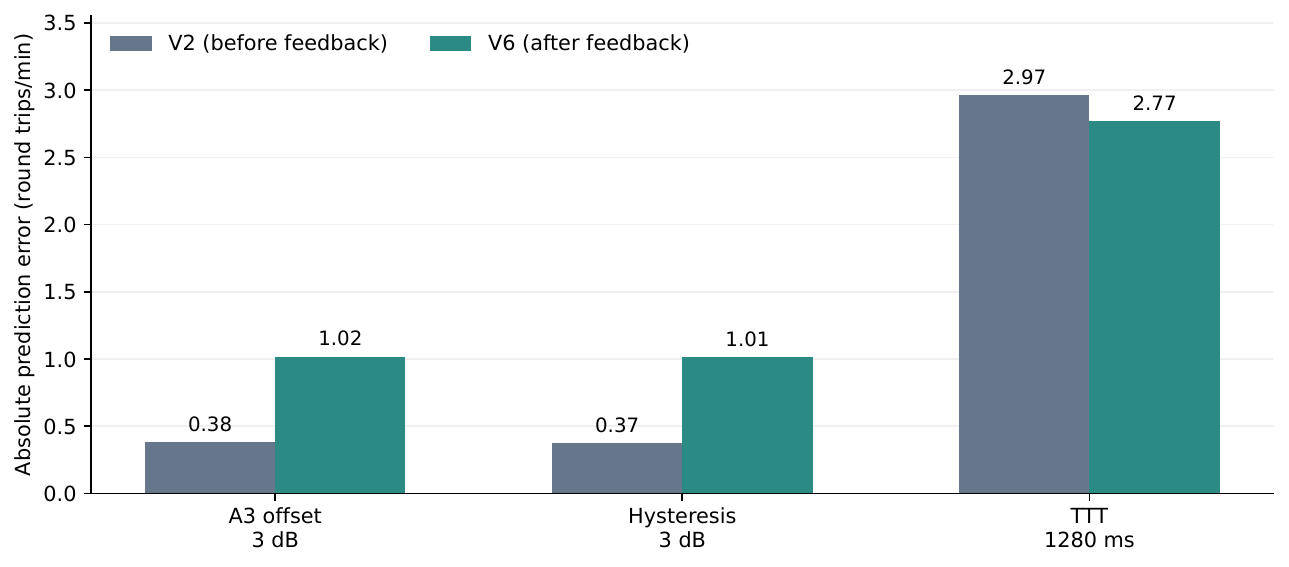}
  \caption{Absolute configuration-effect prediction error before feedback (V2) and after feedback (V6), evaluated against the same reserved physical measurements.}
  \label{fig:error}
\end{figure}

\Cref{tab:prediction,fig:error} compare the two student versions using the same evaluation observations. For TTT $=1280$\,ms, V6 predicts 2.772 rather than V2's 2.966 round trips/min, moving closer to the observed value of zero. For the offset and hysteresis candidates, however, V6 predicts higher frequencies and the absolute errors rise by approximately 0.639 and 0.640 round trips/min. Across all three candidates, MAE changes from 1.240 for V2 to 1.601 round trips/min for V6, an increase of 29.14\%. RMSE changes from 1.740 to 1.803 round trips/min.

The results reveal a parameter-dependent response to feedback. The updated model has the best grouped-validation score among its updated-data candidates, while its errors on new physical parameter magnitudes vary by configuration. The reserved physical block thus supplies a direct diagnosis of which response regions are represented more accurately after adaptation and which remain difficult to predict.

\section{Discussion}
\label{sec:discussion}

\subsection{Learning from Feedback Across Configuration Regions}

The learning and evaluation blocks vary the same parameter types but at different values. The TTT intervention increases from 512\,ms in learning to 1280\,ms in evaluation, and offset and hysteresis increase from 2 to 3\,dB. With only 16 configuration-effect samples distributed across six session groups after the feedback update, the reserved block probes the student's response beyond the magnitudes used for feedback learning.

The contrast between grouped validation and physical evaluation is informative. V6 better fits the historical session-grouped criterion than the alternative updated-data candidates, but its physical prediction errors do not decrease uniformly. In particular, it moves toward the observed TTT response while overpredicting the offset and hysteresis outcomes. A single aggregate validation score conceals this distinction. Conditioning the next experiment on the observed error by parameter is therefore more informative than treating the model update as a uniform improvement.

\subsection{Physical Evidence as Experimental Memory}

A cyber-physical feedback cycle generates more than training examples. The paired interventions identify observed parameter--response relationships; the comparison between predicted and measured outcomes identifies model discrepancies; and the execution records link those discrepancies to realized physical conditions. Stored together, these records provide actionable inputs to the coach.

For example, additional student-oriented measurements could probe TTT values around the duration of the transient attenuation, where event counts change sharply, while sampling offset and hysteresis settings more densely in regions where prediction errors increased. For virtual-environment alignment, the same records would provide targets for checking whether simulated handover events preserve the observed ordering among candidate configurations. This connects physical evidence to the choice of subsequent experiments, rather than using each measurement only once during model training.

\subsection{Physical Execution and Evaluation Scale}

Physical experimentation includes configuration, execution, observation, and restoration. In the case study, all 12 executions retained valid scoring windows and confirmed device-state restoration, showing how the host agent can encapsulate the device operations as auditable experimental transactions. The approximately 16.8\,s windows are suited to observing responses to transient attenuation. As the experimental scope broadens, repeated measurements across mobility and radio conditions, handover failures, radio-link failures, and service-quality outcomes can provide a richer network-level objective for adaptive experiment selection.

\section{Conclusion}
\label{sec:conclusion}

This paper presented Cyber-Physical RSI for AI-RAN, connecting a task-oriented student, an experiment-planning coach, a virtual environment, and a physical host agent. The handover case study implemented an agent-orchestrated physical-feedback pathway comprising 12 executions, six paired interventions, dataset updates, candidate-model selection, and a common reserved physical assessment. The experiments measured different short-window responses to handover parameters, with the longest tested TTT producing no ping-pong round trips in its observed window. The student results showed a lower historical grouped-validation error for the selected updated-data model, alongside configuration-dependent prediction errors on new physical settings. Together, the physical interventions and predictive comparisons demonstrate how measured network behavior can inform both model adaptation and the evidence base for subsequent experimental planning.

\bibliographystyle{unsrtnat}
\bibliography{references}
\end{document}